\pdfoutput=1

\documentclass[11pt]{article}

\usepackage{EMNLP2022}

\usepackage{times}
\usepackage{latexsym}

\usepackage[T1]{fontenc}

\usepackage[utf8]{inputenc}

\usepackage{microtype}

\usepackage{color}
\usepackage{graphicx}
\usepackage{multirow}
\usepackage{makecell}
\usepackage{amsmath}
\usepackage{amsfonts,amssymb}
\usepackage{booktabs}
\usepackage{arydshln}

\title{RASAT: Integrating Relational Structures into Pretrained Seq2Seq Model for Text-to-SQL}
\author{Jiexing Qi\textsuperscript{1},\ Jingyao Tang\textsuperscript{1},\ Ziwei He\textsuperscript{1},\ Xiangpeng Wan\textsuperscript{2},\ Yu Cheng\textsuperscript{3}\\ 
         \textbf{Chenghu Zhou\textsuperscript{4}},\  \textbf{Xinbing Wang\textsuperscript{1}},\ \textbf{Quanshi Zhang\textsuperscript{1}},\ \textbf{Zhouhan Lin\textsuperscript{1}\thanks{\quad  Zhouhan Lin is the corresponding author.}} \\
        \textsuperscript{1}Shanghai Jiao Tong University, Shanghai, China \\
         \textsuperscript{2}NetMind.AI and ProtagoLabs, Virginia, USA \\
        \textsuperscript{3}Microsoft Research, Redmond, Washington, USA \\
        \textsuperscript{4}IGSNRR, Chinese Academy of Sciences, Beijing, China\\
        \texttt{ \{qi\_jiexing, monstar, ziwei.he, zqs1022, xwang8\}@sjtu.edu.cn } \\
        \texttt{lin.zhouhan@gmail.com}}

\begin{document}
\maketitle
\begin{abstract}
Relational structures such as schema linking and schema encoding have been validated as a key component to qualitatively translating natural language into SQL queries. However, introducing these structural relations comes with prices: they often result in a specialized model structure, which largely prohibits using large pretrained models in text-to-SQL. To address this problem, we propose RASAT: a Transformer seq2seq architecture augmented with relation-aware self-attention that could leverage a variety of relational structures while inheriting the pretrained parameters from the T5 model effectively. Our model can incorporate almost all types of existing relations in the literature, and in addition, we propose introducing co-reference relations for the multi-turn scenario. Experimental results on three widely used text-to-SQL datasets, covering both single-turn and multi-turn scenarios, have shown that RASAT could achieve state-of-the-art results across all three benchmarks (75.5\% EX on Spider, 52.6\% IEX on SParC, and 37.4\% IEX on CoSQL). \footnote{Our implementation is available at  \url{https://github.com/LUMIA-group/rasat}. }
\end{abstract}

\section{Introduction}

Text-to-SQL is the task that aims at translating natural language questions into SQL queries. Since it could significantly break down barriers for non-expert users to interact with databases, it is among the most important semantic parsing tasks that are of practical importance \citep{kamath2018survey,deng-etal-2021-structure}. 

Various types of relations have been introduced for this task since \citet{zhong2017seq2sql} collected the first large-scale text-to-SQL dataset, which has resulted in significant boosts in the performance through recent years. For example, \citet{bogin2019representing} introduced schema encoding to represent the schema structure of the database, and the resulting augmented LSTM encoder-decoder architecture was able to generalize better towards unseen database schema. \citet{lin2020bridging} introduced relations between the entity mentioned in the question and the matched entries in the database to utilize database content effectively. Their BERT-based encoder is followed by an LSTM-based pointer network as the decoder, which generalizes better between natural language variations and captures corresponding schema columns more precisely. RAT-SQL \citep{wang-etal-2020-rat} introduced schema linking, which aligns mentions of entity names in the question to the corresponding schema columns or tables. Their augmented Transformer encoder is coupled with a specific tree-decoder. SADGA \citep{NEURIPS2021_3f1656d9} introduced the dependency structure of the natural language question and designed a graph neural network-based encoder with a tree-decoder. On the other hand, a tree-decoder that can generate grammatically correct SQL queries is usually needed to better decode the encoder output, among which  \citet{yin-neubig-2017-syntactic} is one of the most widely used.

Although integrating various relational structures as well as using a tree-decoder have been shown to be vital to generating qualitative SQL queries and generalizing better towards unseen database schema, the dev of various specifically designed model architectures significantly deviate from the general sequential form, which has made it hard if one considers leveraging large pre-trained models for this task. Existing methods either use BERT output as the input embedding of the specifically designed model \citep{cao-etal-2021-lgesql, choi-etal-2021-ryansql, wang-etal-2020-rat, guo-etal-2019-towards}, or stack a specific decoder on top of BERT \citep{lin2020bridging}.

In another thread, pretrained seq2seq models just have unveiled their powerful potential for this task. Recent attempts by \citet{shaw-etal-2021-compositional} show that directly fine-tuning a T5 model \citep{raffel2020t5} on this task without presenting any relational structures could achieve satisfying results. Moreover, PICARD \citep{scholak-etal-2021-picard} presents a way to prune invalid beam search results during inference time, thus drastically improving the grammatical correctness of the SQL queries generated by the autoregressive decoder that comes with T5. 

In this work, different from the more common approach of fine-tuning the original pretrained model or using prompt tuning, we propose to augment the self-attention modules in the encoder and introduce new parameters to the model while still being able to leverage the pre-trained weights. We call the proposed model RASAT\footnote{RASAT: Relation-Aware Self-Attention-augmented T5}.
Our model can incorporate almost all existing types of relations in the literature, including schema encoding, schema linking, syntactic dependency of the question, etc., into a unified relation representation. In addition to that, we also introduce coreference relations to our model for multi-turn text-to-SQL tasks. Experimental results show that RASAT could effectively leverage the advantage of T5. It achieves the state-of-art performance in question execution accuracy (EX/IEX) on both multi-turn (SParC and CoSQL) and single-turn (Spider) text-to-SQL benchmarks.
On SParC, RASAT surpasses all previous methods in interaction execution accuracy (IEX) and improves state-of-the-art performance from 21.6\% to 52.6\%, 31\% absolute improvements. On CoSQL, we improve state-of-the-art IEX performance from 8.4\% to 37.4\%, achieving 29\% absolute improvements. 
Moreover, on Spider, we improve state-of-the-art execution accuracy from 75.1\% to 75.5\%, achieving 0.4\% absolute improvements.

\section{Related Work}
Early works usually exploit a sketch-based slot-filling method that uses different modules to predict the corresponding part of SQL. These methods decompose the SQL generation task into several independent sketches and use different classifiers to predict corresponding part, such as SQLNet \citep{xu2017sqlnet}, SQLOVA \citep{hwang2019comprehensive}, X-SQL \citep{he2019x}, RYANSQL \citep{choi-etal-2021-ryansql}, et.al,.
However, most of these methods only handle simple queries while failing to generate correct SQL in a complex setting such as on Spider. 

Faced with the multi-table and complex SQL setting, using graph structures to encode various complex relationships is a major trend in the text-to-SQL task. 
For example, Global-GNN \citep{bogin-etal-2019-global} represents the complex database schema as a graph,
RAT-SQL \citep{wang-etal-2020-rat} introduces schema encoding and linking and assigns every two input items a relation, LGESQL \citep{cao-etal-2021-lgesql} further distinguishes local and non-local relations by exploiting a line graph enhanced hidden module, SADGA \citep{NEURIPS2021_3f1656d9} uses contextual structure and dependency structure to encode question-graph while database schema relations are used in schema graph, S$^2$SQL \citep{hui2022s} adds syntactic dependency information in relational graph attention network (RGAT) \citep{wang-etal-2020-relational}.

For the conversational context-dependent text-to-SQL task that includes multiple turns of interactions, such as SParC and CoSQL, the key challenge is how to take advantage of historical interaction context.
Edit-SQL \citep{zhang-etal-2019-editing} edits the last turn's predicted SQL to generate the newly predicted SQL at the token level. 
IGSQL \citep{cai-wan-2020-igsql} uses cross-turn and intra-turn schema graph layers to model database schema items in a conversational scenario. 
Tree-SQL \citep{wang2021interactive} uses a tree-structured intermediate representation and assigns a probability to reuse sub-tree of historical Tree-SQLs. 
IST-SQL \citep{wang2021tracking} proposes an interaction state tracking method to predict the SQL query.
RAT-SQL-TC \citep{2112.08735}adds two auxiliary training tasks to explicitly model the semantic changes in both turn grain and conversation grain.
R$^{2}$SQL \citep{hui2021dynamic} and HIE-SQL \citep{zheng2022hie} introduce a dynamic schema-linking graph by adding the current utterance, interaction history utterances, database schema, and the last predicted SQL query.

Recently, \citet{shaw-etal-2021-compositional} showed that fine-tuning a pre-trained T5-3B model could yield results competitive to the then-state-of-the-art. Based on this discovery, \citet{scholak-etal-2021-picard} proposed to constrain the autoregressive decoder through incremental parsing during inference time, effectively filtering out grammatically incorrect sequences on the fly during beam search, which significantly improved the qualities of the generated SQL.

\section{Preliminaries}

\subsection{Task Formulation}
Given a natural language question $\mathcal{Q}$ and database schema $\mathcal{S}= <\mathcal{T}, \mathcal{C}>$, our goal is to predict the SQL query $\mathcal{Y}$. Here $\mathcal{Q}=\{q_i\}^{|\mathcal{Q}|}_{i=1}$ is a sequence of natural language tokens, and the schema $\mathcal{S}$ consists of a series of tables $\mathcal{T}=\{t_i\}^{|\mathcal{T}|}_{i=1}$ with their corresponding columns $\mathcal{C}=\{\mathcal{C}_i\}^{|\mathcal{T}|}_{i=1}$. The content of database $\mathcal{S}$ is noted as $\mathcal{V}$. For each table $t_i$, the columns in this table is denoted as $\mathcal{C}_i=\{c_{ij}\}^{|\mathcal{C}_i|}_{j=1}$. For each table $t_i$, the table name contains $|t_i|$ tokens $t_i={t_{i,1}, \cdots, t_{i,|t_i|}}$, and the same holds for column names. In this work, we present the predicted SQL query as a sequence of tokens, $\mathcal{Y}=\{y_i\}^{|\mathcal{Y}|}_{i=1}$.

In the multi-turn setting, our notations adapt correspondingly. i.e., $\mathcal{Q}=\{\mathcal{Q}_i\}^{|\mathcal{Q}|}_{i=1}$ denotes a sequence of questions in the interaction, with $\mathcal{Q}_i$ denoting each question. Also, the target to be predicted is a sequence of SQL queries, $\mathcal{Y}=\{\mathcal{Y}_i\}^{|\mathcal{Y}|}_{i=1}$, with each $\mathcal{Y}_i$ denoting the corresponding SQL query for the i-th question $\mathcal{Q}_i$. Generally, for each question, there is one corresponding SQL query, such that $|\mathcal{Q}|=|\mathcal{Y}|$. While predicting $\mathcal{Y}_i$, only the questions in the interaction history are available, i.e., $\{\mathcal{Q}_1, \cdots, \mathcal{Q}_{i}\}$. 

\subsection{Relation-aware Self-Attention}
Relation-aware self-attention \citep{shaw-etal-2018-self} augments the vanilla self-attention \citep{NIPS2017_3f5ee243} by introducing relation embeddings into the key and value entries. Assume the input to the self attention is a sequence of $n$ embeddings $X\ =\ \left\{x_{i}\right\}_{i=1}^{n}$ where $x_{i}\in\mathbb{R}^{d_{x}}$, then it calculates its output $z$ as (\ $||$ means concatenate operation):
\begin{equation}
\small
\label{eq:self-attention}
\begin{aligned}
\alpha_{i j}^{(h)} &= \operatorname{softmax}\left( \frac{\boldsymbol{x}_{\boldsymbol{i}} W_{Q}^{(h)}\left(\boldsymbol{x}_{\boldsymbol{j}} W_{K}^{(h)} + \boldsymbol{r}_{i j}^{K} \right)^{\top}}{\sqrt{d_{z} / H}} \right)\\
\boldsymbol{z}_{i} &= \mathop{\bigg|\bigg|}\limits_{h=1}^{H} \left[\ \sum_{j=1}^{n} \alpha_{i j}^{(h)}\left(\boldsymbol{x}_{\boldsymbol{j}} W_{V}^{(h)} +\boldsymbol{r}_{i j}^{V} \right)\ \right]\\
\end{aligned}
\end{equation}

where  $H$ is the number of heads, and $W_{Q}^{(h)},W_{K}^{(h)},W_{V}^{(h)}$ are learnable weights. The $\boldsymbol{r}_{i j}^{K}, \boldsymbol{r}_{i j}^{V}$ are two different relation embeddings used to represent the relation $r$ between the $i$-th and $j$-th token.  

\begin{figure*}[t]
    \centering
    \includegraphics[width=2.0\columnwidth]{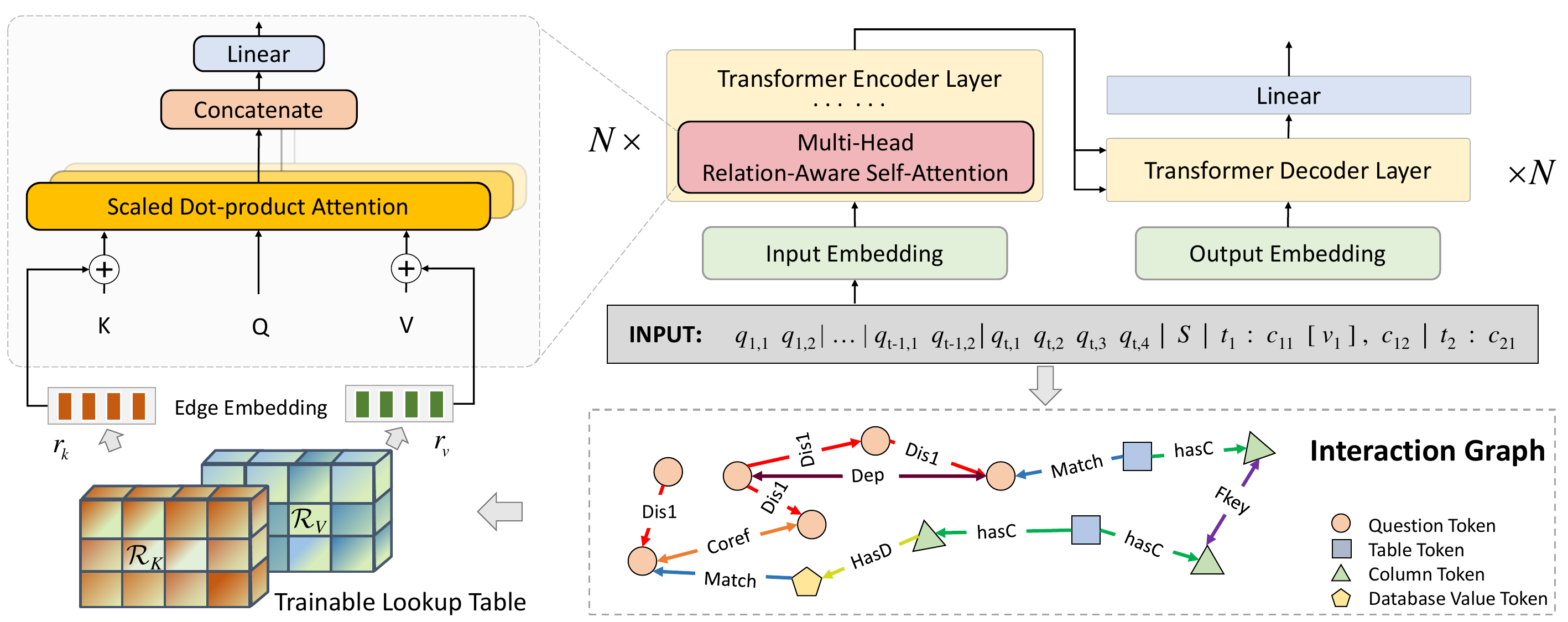}
    \caption{The overview of our model. Our model inherits the seq2seq architecture of T5, consisting of $N$ layers of encoders and decoders. The self-attention modules in the encoder are substituted with relation-aware self-attention, introducing two additional relation embedding lookup tables $\mathcal{R}_K$ and $\mathcal{R}_V$. We convert the sequential input into an interaction graph by introducing various types of relations and adapting them to the subword level through relation propagation. 
    During the forward process, the relation-aware self-attention modules read out the relations of each token through the interaction graph and retrieve the corresponding relations embeddings from the lookup tables $\mathcal{R}_K$ and $\mathcal{R}_V$. }
    \label{fig:model}
\end{figure*}

\begin{table*}[!ht]
\centering
\resizebox{2.0\columnwidth}{!}{
\begin{tabular}{lccll}
\toprule
\textbf{Type }                            & \textbf{Head   H}         & \textbf{Tail   T}           & \textbf{Edge   Label}  & \textbf{Description}                                                  \\ \midrule
\multirow{3}{*}{Schema Encoding} & \multirow{2}{*}{$\mathcal{T}$} & \multirow{2}{*}{$\mathcal{C}$}   & \textsc{Primary-Key}   & T is the primary-key for H                                   \\
                                 &                    &                      & \textsc{Belongs-to}    & T is a column in H                                           \\ \cmidrule{2-5} 
                                 & $\mathcal{C}$                  & $\mathcal{C}$                    & \textsc{Foreign-Key}   & H is the foreign key for T                                   \\ \midrule
\multirow{2}{*}{Schema Linking}  & \multirow{2}{*}{$\mathcal{Q}$} & \multirow{2}{*}{$\mathcal{T}$/$\mathcal{C}$} & \textsc{Exact-Match}   & H is part of T, and T is a span of the entire question     \\
                                 &                    &                      & \textsc{Partial-Match} & H is part of T, but the entire   question does not contain T \\ \midrule
Question Dependency              & $\mathcal{Q}$                  & $\mathcal{Q}$                    & \textsc{Dependency}    & H has a forward syntactic dependencies   on T                \\ \midrule
Question Coreference             & $\mathcal{Q}$                  & $\mathcal{Q}$                    & \textsc{Coreference}   & H is the coreference of T                                    \\ \midrule
Database Content                 & $\mathcal{Q}$                  & $\mathcal{C}$                    & \textsc{Value-Match}   & H is part of the candidate cell values   of column T         \\ 
\bottomrule
\end{tabular}}
\caption{Description of some representatives for each relation type in the interaction graph. For a complete list of relations, please refer to Appendix \ref{sec:appendix_relation}.}
\label{table:relation}
\end{table*}

\section{RASAT}
\subsection{Model Overview}
The overall structure of our RASAT model is shown in Figure \ref{fig:model}. Architecture-wise it is rather simple: the T5 model is taken as the base model, with its self-attention modules in the encoder substituted as relation-aware self-attentions. 

The input to the encoder is a combination of question(s) $\mathcal{Q}$, database schema $\mathcal{S}=<\mathcal{T}, \mathcal{C}>$ with the database name $S$, as well as database content mentions and necessary delimiters. We mostly follow  \citet{shaw-etal-2021-compositional} and  \citet{scholak-etal-2021-picard} to serialize the inputs. Formally, 
\begin{equation}
X= \overline{\mathcal{Q} \texttt{|} S \texttt{|} t_{1} \texttt{:} c_{11} \texttt{[} v \texttt{]} \texttt{,} \cdots \texttt{,} c_{1|T_1|} \texttt{|}t_{2} \texttt{:} c_{21} \texttt{,} \cdots}
\end{equation}

where $t_i$ is the table name, $c_{ij}$ is the $j$-th column name of the $i$-th table. The $v \in \mathcal{V}$ showing after column $c_{11}$ is the database content belonging to the column that has n-gram matches with the tokens in the question. As for delimiters, we use $\texttt{|}$ to note the boundaries between $\mathcal{Q}$, $S$, and different tables in the schema. Within each table, we use $\texttt{:}$ to separate between table name and its columns. Between each column, $\texttt{,}$ is used as the delimiter. 

As for the multi-turn scenario, we add the questions in the history at the start of the sequence and truncate the trailing tokens in the front of the sequence when the sequence length reaches 512. i.e.,
\begin{equation}
X= \overline{\mathcal{Q}_{1} \texttt{|} \mathcal{Q}_{2} \texttt{|} \cdots \texttt{|} \mathcal{Q}_t \texttt{|} S \texttt{|} t_{1} \texttt{:} c_{11} \texttt{[} v \texttt{]} \texttt{,} \cdots}
\end{equation}

where $\texttt{|}$ are the corresponding delimiters. 

Next, we add various types of relations as triplets, linking between tokens in the serialized input, which naturally turns the input sequence into a graph (Figure \ref{fig:model}). We will elaborate on this in Subsection \ref{interaction_graph}. Moreover, since almost all relation triplets, its head and tail correspond to either a word or a phrase, while the T5 model is at subword level, we also introduce relation propagation to map these relations to subword level, which is detailed in Subsection \ref{relation_propagation}.

To fine-tune this model, we inherit all the parameters from T5 and randomly initialize the extra relation embeddings introduced by relation-aware self-attention. 
The overall increase of parameters is less than 0.01\% (c.f. Appendix \ref{sec:appendix_paramerter}).

\subsection{Interaction Graph} \label{interaction_graph}
Equipped with relation-aware self-attention, we can incorporate various types of relations into the T5 model, as long as the relation can be presented as a triplet, with its head and tail being the tokens in the input sequence $X$. Formally, we present the triplet as 
\begin{equation}
<H, r, T>
\end{equation}
where $H, T$ are the head and tail items in the triplet, and $r$ represents the relation.

Given the input sequence $X$ of length $|X|$, we assume that for each direction of a given pair of tokens, there only exists up to one relation. Thus, if we consider the tokens in $X$ as vertices of a graph, it could have up to ${|X|}^2$ directed edges, with each edge corresponding to an entry in the adjacency matrix of the graph. In this paper, we call this graph, containing tokens from the whole input sequence as its vertices and the incorporated relations as its edges, as \emph{interaction graph}.

We assign two relation embeddings for each type of introduced relation. Thus the Transformer encoder comes with two trainable lookup tables storing relations embeddings to compute the key and value in the self-attention (c.f. Figure \ref{fig:model}). Formally, we denote them as $\mathcal{R}_K,\mathcal{R}_V \in \mathbb{R}^{\mu\times d_{kv}}$ where $\mu$ is the kinds of relations and $d_{kv}$ is the dimension of each attention head in the key and value states. Note that we share the relation embedding between different heads and layers but untie them between key and value. During forward computation, for all the layers, $r_{ij}^K$ and $r_{ij}^V$ in Equation \ref{eq:self-attention} are retrieved from the two trainable look-up tables.

We reserve a set of \emph{generic} relations for serving as mock relations for token pairs that do not have a specific edge. In total, we have used 51 different relations in the model (c.f. Appendix \ref{sec:appendix_relation}). Apart from the mock \emph{generic} relations, there are generally 5 types of relations, which are: \emph{schema encoding, schema linking, question dependency structure, coreference between questions,} and \emph{database content mentions}. Please refer to Table \ref{table:relation} for some representative examples for each type. We will describe each of them in the following paragraphs. 

\paragraph{Schema Encoding.} Schema encoding relations refer to the relation between schema items, i.e., $H, T \in \mathcal{S}$. These relations describe the structure information in a database schema. For example, \textsc{Primary-Key} indicates which column is the primary key of a table, \textsc{Belongs-To} shows which table a column belongs to, and \textsc{Forign-Key} connects the foreign key in one table, and the primary key in another table.
    
\paragraph{Schema Linking.} Schema linking relations refer to the relations between schema and question items, i.e., $H \in \mathcal{S}, T \in  \mathcal{Q}$ or vice versa. We follow the settings in RAT-SQL \citep{wang-etal-2020-rat}, which uses n-gram matches to indicate question mentions of the schema items. Detecting these relations is shown to be challenging in previous works \citep{guo-etal-2019-towards,deng-etal-2021-structure} due to the common mismatch between natural language references and their actual names in the schema. Thus, we also discriminate between exact matches and partial matches to suppress the noise caused by imperfect matches.

\begin{figure}
    \centering
    \includegraphics[width=1.0\columnwidth]{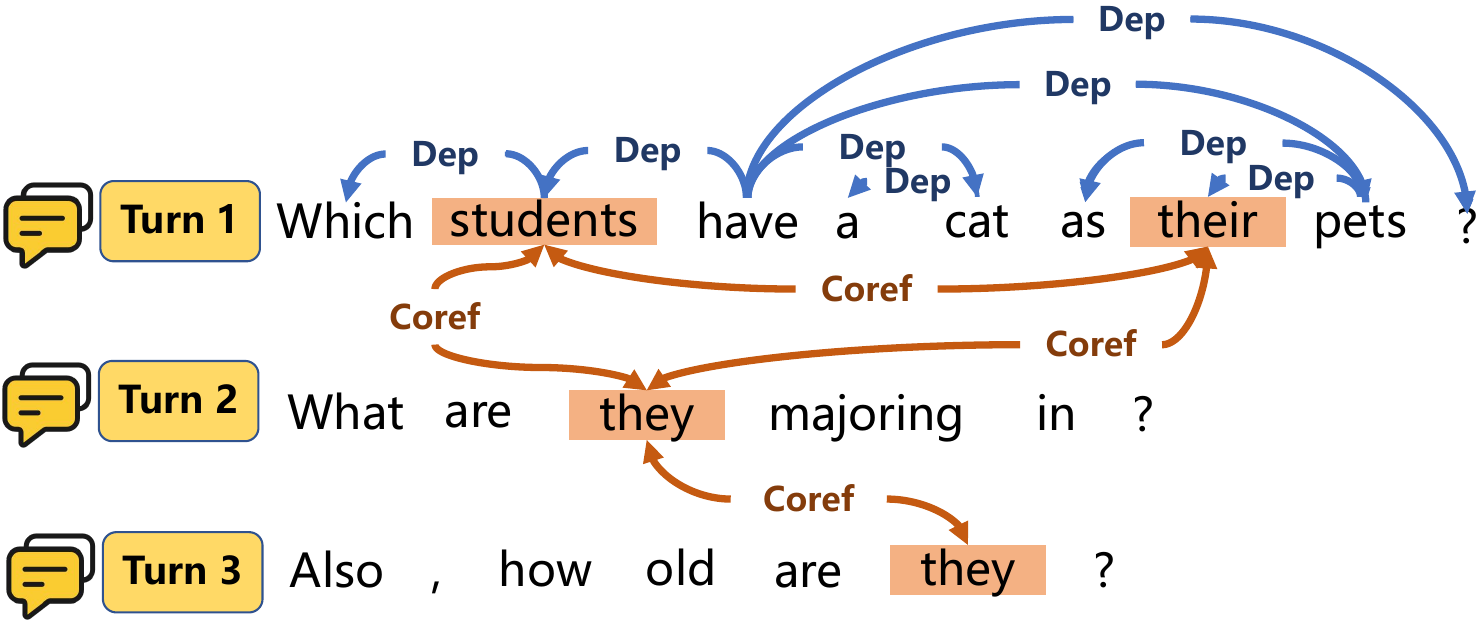}
    \caption{An example to show the coreference and syntactic dependency relations on user questions between different turns.}
    \label{fig:coref}
\end{figure}

    
\paragraph{Question Dependency Structure.} This type of relation refers to the edges of a dependency tree of the question, i.e., $H, T \in \mathcal{Q}$. Unlike the previous two relation types, it is less explored in the literature on text-to-SQL. Since it reflects the grammatical structure of the question, we believe it should also be beneficial for the task. In our work, to control the total number of relations and avoid unnecessary overfitting, we do not discriminate between different dependency relations. Figure \ref{fig:coref} shows an example of dependency relations in one of its questions. 

\paragraph{Coreference Between Questions.} This type of relation is unique to the multi-turn scenario. 
In a dialog with multiple turns, it is important for the model to figure out the referent of the pronouns correctly. Figure \ref{fig:coref} shows a typical case of coreference resolution. The question item "their" in Turn 1, "they" in Turn 2, and "they" in Turn 3 all refer to the question item "students" in Turn 1. i.e., $H \in \mathcal{Q}_i, T \in \mathcal{Q}_j$. To our best knowledge, there are no works utilizing this relation in the text-to-SQL literature despite the importance of this relation. Although pre-trained models like T5 are believed to have the capability to handle this implicitly, we still find that explicitly adding these links could significantly improve the model's performance.

\paragraph{Database Content Mentions.} Instead of mentioning the table or column names, the user could mention the values in a specific column. In this case, the informative mention could escape from the aforementioned schema linking. In this work, we follow the same procedures in BRIDGE \citep{lin-etal-2020-bridging} to capture database content mentions. It first performs a fuzzy string match between the question tokens and the values of each column in the database. i.e., $H \in \mathcal{Q}, T\in\mathcal{V}$. Then the matched values are inserted after the corresponding column name in the input sequence. This relation is denoted as \textsc{Value-Match} in Table \ref{table:relation} and is also widely used in many graph-structured models \citep{wang-etal-2020-rat, cao-etal-2021-lgesql}.

\begin{table*}[t]
\centering
\resizebox{2.0\columnwidth}{!}{
\begin{tabular}{lccccccccc}
\toprule
\multirow{2}{*}{Approach} & \multicolumn{4}{c}{Dev}   &  & \multicolumn{4}{c}{Test}  \\ \cmidrule{2-5} \cmidrule{7-10} 
                          & QEM  & IEM  & QEX  & IEX  &  & QEM  & IEM  & QEX  & IEX  \\ \midrule
RAT-SQL + SCoRe \citep{yu2020score}          & 62.2 & 42.5 & -    & -    &  & 62.4 & 38.1 &  -   &   -  \\
HIE-SQL + GraPPa \citep{zheng2022hie}         & 64.7 & 45.0 & -    & -    &  & 64.6 & 42.9 & -    & -    \\
RAT-SQL-TC + GAP \citep{2112.08735}         & 64.1 & 44.1 & -    & -    &  & 65.7 & 43.2 & -    & -    \\ \midrule
GAZP+BERT  \citep{zhong-etal-2020-grounded}               & 48.9 & 29.7 & 47.8 & -    &  & 45.9 & 23.5 & 44.6 & 19.7 \\
TreeSQL v2+BERT \citep{wang2021interactive}          & 52.6 & 34.4 & 50.4 & 29.4 &  & 48.1 & 25.0 & 48.5 & 21.6 \\
UNIFIEDSKG  \citep{xie2022unifiedskg}              & 61.5 & 41.9 & 67.3 & 46.4 &  & -    & -    & -    & -    \\ \midrule
RASAT                     & 65.0 & 45.7 & 69.9 & 50.7 &  & -    & -    & -    & -    \\
RASAT+PICARD              & \textbf{67.7} & \textbf{49.1} & \textbf{73.3} & \textbf{54.0} &  & \textbf{67.7}  & \textbf{45.2}    & \textbf{74.0}    & \textbf{52.6}    \\ \bottomrule
\end{tabular}}
\caption{Results on SParC dataset. Models in the upper block do not predict SQL values, while the ones in the middle block do.}
\label{tab:sparc_dev_res}
\end{table*}

\subsection{Relation Propagation}    \label{relation_propagation}
The various aforementioned types of relations are between types of items, with their $H$ and $T$ being either words or phrases. However, almost all pre-trained models take input tokens at the subword level, resulting in a difference in the granularity between the relations and the input tokens. Previous works use an extra step to aggregate multiple subword tokens to obtain a single embedding for each item in the interaction graph, such as mean pooling, attentive pooling, or with BiLSTMs \citep{wang-etal-2020-rat,cao-etal-2021-lgesql}. However, these aggregation methods are detrimental to inheriting the pre-trained knowledge in the pretrained models.

In this work, we adopt the other way: we propagate the relations into the subword level by creating a dense connection of the same type of relations between the tokens in $H$ and $T$. For example, column \texttt{amenid} is a foreign key in table \texttt{has\_amenity} and the corresponding primary key is column \texttt{amenid} in table \texttt{dorm\_amenity}. Such that there is a directed relation \textsc{Foreign-Key} between the two column names. At subword level, \texttt{amenid} consists of two tokens \texttt{amen} and \texttt{id}. Accordingly, we propagate the \textsc{Foreign-Key} relation into 4 replicas, pointing from tokens in the source \texttt{amenid} to that of the target one, forming a dense connection between subword tokens on both sides. With relation propagating, we could conveniently adapt word or phrase level relations to our RASAT model while keeping the pretrained weights learned at the subword level intact.

\begin{table}
\centering
\resizebox{1\columnwidth}{!}{
\begin{tabular}{lllll}
\toprule
      & \textit{Spider} & \textit{Realistic} & \textit{SParC}     & \textit{CoSQL}     \\ 
\midrule
Train & 7000   & -             & 3034/9025 & 2164/7485 \\
Dev   & 1034   &  508             & 422/1203  & 292/1008  \\
Test  & 2147   & -         & 842/2498  & 551/2546     \\ 
\bottomrule
\end{tabular}}
\caption{Dataset statistics for Spider, Spider-Realistic (\textit{Realistic} in table), SParC and CoSQL. For Spider and Spider-Realistic, the table shows the number of question-SQL pairs in the train-dev-test splits. For SParC and CoSQL, we list both the number of interactions and questions in the form of "\#interactions/\#questions".}
\label{tab:stat_for_datasets}
\end{table}

\begin{table*}
\centering
\resizebox{2.0\columnwidth}{!}{
\begin{tabular}{lccccccccc}
\toprule
\multirow{2}{*}{Approach} & \multicolumn{4}{c}{Dev}                                       &  & \multicolumn{4}{c}{Test}                          \\ \cmidrule{2-5} \cmidrule{7-10} 
                          & QEM           & IEM           & QEX           & IEX           &  & QEM        & IEM        & QEX        & IEX        \\ \midrule
RAT-SQL + SCoRe \citep{yu2020score}          & 52.1          & 22.0          & -             & -             &  & 51.6       & 21.2       & -          & -          \\
HIE-SQL + GraPPa \citep{zheng2022hie}         & 56.4          & \textbf{28.7}          & -             & -             &  & 53.9       & 24.6       & -          & -          \\ \midrule
GAZP+BERT \citep{zhong-etal-2020-grounded}                 & 42.0          & 12.3          & 38.8          & -             &  & 39.7       & 12.8       & 35.9       & 8.4          \\ 
UNIFIEDSKG \citep{xie2022unifiedskg}               & 54.1          & 22.8          & 62.2          & 26.2          &  & -          & -          & -          & -          \\
T5-3B \citep{scholak-etal-2021-picard}                    & 53.8          & 21.8          & -             & -             &  & 51.4       & 21.7       & -          & -          \\
T5-3B+PICARD \citep{scholak-etal-2021-picard}             & 56.9          & 24.2          & -             & -             &  & 54.6       & 23.7       & -          & -          \\ \midrule
RASAT                     & 56.2          & 25.9          & 63.8          & 34.8          &  & -          & -          & -          & -          \\
RASAT+PICARD              & \textbf{58.8} & 27.0 & \textbf{67.0} & \textbf{39.6} &  & \textbf{55.7} & \textbf{26.5} & \textbf{66.3} & \textbf{37.4} \\ \bottomrule
\end{tabular}}
\caption{Results on CoSQL dataset. Models in the upper block do not predict SQL values, while the ones in the middle block do.}
\label{tab:cosql_dev_res}
\end{table*}

\begin{table}[ht]
\resizebox{1\columnwidth}{!}{
\begin{tabular}{lcccc}
\toprule
\multirow{2}{*}{Approach} & \multicolumn{2}{c}{Dev} & \multicolumn{2}{c}{Test} \\ \cmidrule{2-5} 
                          & EM         & EX         & EM          & EX         \\ \midrule
RAT-SQL+BERT              & 69.7       & -          & 65.6        & -          \\
LGESQL+ELECTRA            & 75.1       & -          & 72.0        & -          \\
S²SQL   + ELECTRA         & \textbf{76.4} & -       & \textbf{72.1} &  -       \\ \midrule
BRIDGE v2+BERT            & 71.1       & 70.3       & 67.5        & 68.3       \\
NatSQL+GAP                & 73.7       & 75.0       & 68.7        & 73.3       \\
SmBoP   + GraPPa          & 74.7       & 75.0       & 69.5        & 71.1       \\
T5-3B                     & 71.5       & 74.4       & 68.0        & 70.1       \\
T5-3B + PICARD            & 75.5       & 79.3       & 71.9        & 75.1       \\ \midrule
RASAT                     & 72.6       & 76.6       & -           & -          \\
RASAT+PICARD              & 75.3       & \textbf{80.5} & 70.9    & \textbf{75.5}          \\ \bottomrule
\end{tabular}}
\caption{Results on Spider dataset. Models in the upper block do not predict SQL values, while the ones in the middle block do. We compare RASAT with some important baseline methods, such as RAT-SQL \citep{wang-etal-2020-rat}, Bridge \citep{lin-etal-2020-bridging}, GAZP \citep{zhong-etal-2020-grounded}, NatSQL \citep{gan-etal-2021-natural-sql}, SmBoP \citep{rubin-berant-2021-smbop}, LGESQL \citep{cao-etal-2021-lgesql}, S$^2$SQL \citep{hui2022s},  T5 and PICARD \citep{scholak-etal-2021-picard}. }
\label{tab:spider}
\end{table}

\begin{table}
\resizebox{1\columnwidth}{!}{
\begin{tabular}{lcc}
\toprule
Approach      & EM   & EX   \\ \midrule
RAT-SQL+STRUG \citep{deng-etal-2021-structure}  & 62.2 & 65.7 \\
T5-3B  \citep{scholak-etal-2021-picard}       & 62.0 & 64.1 \\
T5-3B+PICARD \citep{scholak-etal-2021-picard} & 68.7 & 71.4  \\ \midrule
RASAT         & 65.2 & 65.8 \\
RASAT+PICARD  & \textbf{69.7} & \textbf{71.9} \\ \bottomrule
\end{tabular}}\caption{Results on Spider-Realistic dataset. We reproduce \citet{scholak-etal-2021-picard} 's method to get the performance of T5-3B (+PICARD) and the performance of RAT-SQL+STRUG are from \citet{deng-etal-2021-structure} reported.}
\label{tab:spider_realistic}
\end{table}

\section{Experiments}
In this section, we will show our model's performance on three common text-to-SQL datasets: Spider \citep{yu-etal-2018-spider}, SParC \citep{yu-etal-2019-sparc} and CoSQL \citep{yu-etal-2019-cosql}. Besides, we experiment on a more realistic setting of the Spider dataset: Spider-Realistic \citep{deng-etal-2021-structure} to test the generalizability of our model. The statistics of these datasets are shown in Table \ref{tab:stat_for_datasets}. We also present a set of ablation studies to show the effect of our method on different sized models, as well as the relative contribution of different relations. In addition, we put 2 case studies in Appendix \ref{sec:appendix_case_study}.

\subsection{Experiment Setup}

\paragraph{Datasets} Spider is a large-scale, multi-domain, and cross-database benchmark. SparC and CoSQL are multi-turn versions of Spider on which the dialogue state tracking is required. All test data is hidden to ensure fairness, and we submit our model to the organizer of the challenge for evaluation.

\paragraph{Evaluation Metrics}
We use the official evaluation metrics: Exact Match accuracy (EM) and EXecution accuracy (EX). EM measures whether the whole predicted sequence is equivalent to the ground truth SQL (without values), while in EX, it measures if the predicted executable SQLs (with values) can produce the same result as the corresponding gold SQLs. As for SParC and CoSQL, which involve a multi-turn scenario, both EM and EX can be measured at the question and interaction levels. Thus we have four evaluation metrics for these two datasets, namely Question-level Exact Match (QEM), Interaction-level Exact Match (IEM), question-level EXecution accuracy (QEX), and interaction-level EXecution accuracy (IEX).

\paragraph{Implementation}  Our code is based on Huggingface transformers \citep{wolf-etal-2020-transformers}. 
We align most of the hyperparameter settings with \citet{shaw-etal-2021-compositional} to provide a fair comparison. 
For coreference resolution, we use coreferee\footnote{https://github.com/msg-systems/coreferee} to yield coreference links. In total, 51 types of relations are used (c.f. Appendix \ref{sec:appendix_relation} for a detailed list). For dependency parsing, stanza \citep{qi-etal-2020-stanza} is used. The batch size we used is 2048. We use Adafactor \citep{shazeer2018adafactor} as optimizer and the learning rate is 1e-4. We set "parse with guards" mode for PICARD and beam size is set to 8. The max tokens to check for PICARD is 2. Experiments are run on NVIDIA A100-SXM4-80GB GPUs.

\subsection{Results on SParC}

The results on SParC are shown in Table \ref{tab:sparc_dev_res}. Our proposed RASAT model combined with PICARD achieves state-of-the-art results on all four evaluation metrics.

Compared with the previous state-of-the-art RAT-SQL-TC + GAP \citep{2112.08735}, RASAT + PICARD brings the QEM from 65.7\% to 67.7\% and IEM from 43.2\% to 45.2\% on the test set. In addition, our model can produce executable SQLs (with values), whereas many of the models listed in the table do not provide value predictions. 

Among the models that can predict with values, the fine-tuned T5-3B model from UNIFIEDSKG \citep{xie2022unifiedskg} is currently the state-of-the-art. We did comparison of QEX/IEX on the dev set since they did not report their performance on the test set.
RASAT + PICARD surpasses all previous methods and improves the state-of-art QEX and IEX from 67.3\% and 46.4\% to 73.3\% and 54.0\%, with 6\% and 7.6\% absolute improvements, respectively. 

Furthermore, on the official leaderboard of SParc which reports over test set, our proposed RASAT + PICARD brings the IEX from 21.6\% to 52.6\%, achieving 31\% absolute improvements.


\subsection{Results on CoSQL}
Compared with SParC, CoSQL is labeled in a Wizard-of-Oz fashion, forming a more realistic and challenging testbed. Nevertheless, our proposed model could still achieve state-of-the-art results (Table \ref{tab:cosql_dev_res}) on all four evaluation metrics.



By comparing to the previous state-of-the-art HIE-SQL + GraPPa \citep{zheng2022hie} and T5-3B+PICARD \cite{scholak-etal-2021-picard}, RASAT + PICARD brings the QEM from 54.6\% to 55.7\% and IEM from 24.6\% to 26.5\% on the test set.  

For the same reason as on SParC, we mainly compare QEX/IEX performance on the dev set, and RASAT + PICARD surpasses all models that can predict executable SQLs (with values). Especially for IEX, our model surpasses the previous state-of-the-art from 26.2\% to 39.6\%, with 13.4\%  absolute improvement. 
Moreover, on the official leaderboard of CoSQL which reports over test set, RASAT + PICARD brings the IEX from 8.4\% to 37.4\%, with 29\% absolute improvements.

\subsection{Results on Spider and Spider-Realistic}
The results on the Spider is provided in Table \ref{tab:spider}. Our proposed RASAT model achieves state-of-the-art performance in EX and competitive results in EM. 
On the dev set, compared with T5-3B, which also does not use the PICARD during beam search, our model's EX increases from 74.4\% to 76.6\%, achieving 2.2\% absolute improvement. When augmented with PICARD, RASAT+PICARD brings the EX even higher to 80.5\%, with 1.2\% absolute improvement compared to T5-3B + PICARD. 
Furthermore, on the official leaderboard of Spider, our proposed RASAT + PICARD brings the EX from 75.1\% to 75.5\%, achieving new state-of-the-art.

Furthermore, we also evaluate our model on a more challenging Spider variant, Spider-Realistic \citep{deng-etal-2021-structure}. It is a evaluation dataset that has modified the user questions by removing or paraphrasing explicit mentions of column names to present a realistic and challenging setting. Our model also achieves a new state-of-the-art performance (Table \ref{tab:spider_realistic}), which suggests strong ability of our model to generalize to unseen data.

\subsection{Ablation Study}
In this subsection, we conduct a set of ablation studies to examine various aspects of the proposed model. Due to the limited availability of the test sets, all numbers in this subsection are reported on the dev set. 
\paragraph{Effect on SQL difficulty.} One might conjecture that the introduced relations are only effective for more difficult, longer SQL query predictions, while for predicting short SQL queries, the original T5 model could handle equally well. Thus, we evaluate our model according to the difficulty of the examples, where the question/SQL pairs in the dev set are categorized into four subsets, i.e., easy, medium, hard, and extra hard, according to their level of difficulty. In Table \ref{tab:spider-group} we provide a comparison between T5-3B + PICARD \citep{scholak-etal-2021-picard} and RASAT + PICARD on the EX metric on the four subsets. RASAT + PICARD surpasses T5-3B + PICARD across all subsets, validating the effectiveness of the introduced relational structures for all SQL sequences. 

\begin{table}
\resizebox{1\columnwidth}{!}{
\begin{tabular}{lcccc}
\toprule
Approach     & easy & medium & hard & extra  \\ 
\midrule
T5-3B + PICARD & 95.2 & 85.4   & 67.2 & 50.6       \\
RASAT + PICARD & \textbf{96.0} & \textbf{86.5}   & \textbf{67.8} & \textbf{53.6}       \\ 
\bottomrule
\end{tabular}}
\caption{EX accuracy of RASAT+PICARD and T5-3B+PICARD on the examples of Spider dev set with different levels of difficulty.}
\label{tab:spider-group}
\end{table}

\begin{table}
\small
\centering
\begin{tabular}{lll}
\toprule
Approach      & EM                  & EX                  \\ 
\midrule
T5-small      & 47.2                & 47.8                \\
RASAT(-small) & \textbf{53.0(+5.8)} & \textbf{53.7(+5.9)} \\ 
\midrule
T5-base       & 57.2                & 57.9                \\
RASAT(-base)  & \textbf{60.4(+3.2)} & \textbf{61.3(+3.4)} \\ 
\midrule
T5-large      & 65.3                & 67.2                \\
RASAT(-large) & \textbf{66.7(+1.4)} & \textbf{69.2(+2.0)} \\ 
\midrule
T5-3B         & 71.5                & 74.4                \\
RASAT(-3B)    & \textbf{72.6(+1.1)} & \textbf{76.6(+2.2)} \\ 
\bottomrule
\end{tabular}
\caption{Result for different T5 model sizes on Spider dev set. The performance of T5 baselines are from \citet{scholak-etal-2021-picard}.}
\label{tab:ablation_size}
\end{table}

\begin{table}
\centering
\small
\begin{tabular}{lll}
\toprule
Approach        & EM         & EX         \\ 
\midrule
T5-small        & 47.2       & 47.8       \\
\quad w/o db\_content & 45.8(-1.4) & 46.9(-0.9) \\
\midrule
RASAT(-small)    & \textbf{53.0}       & \textbf{53.7}       \\
\quad w/o db\_content & 52.6(-0.4) & 52.9(-0.8) \\
\quad w/o dependency    & 51.3(-1.7) & 51.7(-2.0) \\
\bottomrule
\end{tabular}
\caption{Ablation study on the relative contribution of different relation types. Experiment are conducted using RASAT(-small) on the Spider dataset.}
\label{tab:spider_relation_ablation}
\end{table}

\begin{table}
\resizebox{1\columnwidth}{!}{
\begin{tabular}{lllll}
\toprule
Approach        & QEM               & IEM              & QEX                 & IEX                 \\ 
\midrule
RASAT           & 64.5              & \textbf{45.7}    & 69.2                & 50.4                \\
\quad w/o Dp    & \textbf{65.0(+0.5)}& 45.5(-0.2)   & \textbf{69.9(+0.7)} & \textbf{50.7(+0.3)} \\
\quad w/o Cf    & \textbf{65.0(+0.5)}& 45.0(-0.7)     & 69.4(+0.2)          & 50.0(-0.4)            \\
\quad w/o Db    & 64.1(-0.4)        & 45.3(-0.4)       & 67.9(-1.3)          & 48.5(-1.9)          \\
\quad w/o SL    & 64.5              & 45.5(-0.2)       & 68.8(-0.4)          & 49.4(-1.0)          \\
\quad w/o SE    & 63.9(-0.6)        & 44.6(-1.1)       & 68.6(-0.6)          & 48.9(-1.5)          \\
\bottomrule
\end{tabular}}
\caption{Ablation study on the relative contribution of different relation types. Experiment are conducted using RASAT(-3B) on the SParC dataset. ``Dp'' is short for dependency relation, ``Cf'' for coreference relation, ``SL'' for schema linking relation, ``SE'' for schema encoding relation and ``Db'' means database content.}
\label{tab:sparc_relation_ablation}
\end{table}

\paragraph{Model Size Impact.} 
To test the effectiveness of the introduced relational structures on pretrained models with different sizes, we implant RASAT into four T5 models of different sizes (T5-small, T5-base, T5-large, T5-3B) and test it on Spider (Table \ref{tab:ablation_size}). Interestingly, for smaller pretrained models, our RASAT model could bring even larger performance gaps between its T5-3B counterpart. This suggests that the larger T5 model might have learned some of the relational structures implicitly. We believe this is consistent with the findings on other fine-tuning tasks, where larger pretrained models are more capable of capturing the abundant implicit dependencies in the raw text.



\paragraph{Relation Types.} 
We conducted additional experiments to analyze the relative contribution of different relation types. The experimental results on Spider is shown in Table \ref{tab:spider_relation_ablation} while result on SParC is shown in Table \ref{tab:sparc_relation_ablation} (since CoSQL has similar conversational modality with SParC, the experiments are only conducted on SParC). We find that both T5 and RASAT models can benefit from leveraging database content.
Another important finding is that the performance has increased obviously by adding dependency relationship to RASAT(-small) on Spider. 
As for SParC, the database content plays a more important role by looking at EX results; from what we can see, IEX will decrease by 1.9\% after removing database content from the input.



\section{Conclusion}
In this work, we propose RASAT, a Relation-Aware Self-Attention-augmented T5 model for the text-to-SQL generation. Compared with previous work, RASAT can introduce various structural relations into the sequential T5 model. Different from the more common approach of fine-tuning the original model or using prompt tuning, we propose to augment the self-attention modules in the encoder and introduce new parameters to the model while still being able to leverage the pre-trained weights. RASAT had achieved state-of-the-art performances, especially on execution accuracy, in the three most common text-to-SQL benchmarks.


\section*{Limitation}
Our method consumes plenty of computational resources since we leverage the large T5-3B model. We train our models on 8 A100 GPUs (80G) for around 2 days.
Our model truncates the source sequences to 512, this may lead to information loss when a sample has long input. We find that about 3\% of training data in CoSQL will be affected.
We only work with English since it has richer analytical tools and resources than other language. 

\section*{Acknowledgement}
This work was sponsored by the National Natural Science Foundation of China (NSFC) grant (No.
62106143), and Shanghai Pujiang Program (No. 21PJ1405700). We would like to thank Tao Yu, Hongjin Su, and Yusen Zhang for running evaluations on our submitted models. 
\clearpage

\bibliography{anthology,custom}
\bibliographystyle{acl_natbib}

\newpage

\appendix
\clearpage

\section{Model Size}
\label{sec:appendix_paramerter}
Compared with the original T5 model, only two embedding matrices are added to the encoder in our model, with $2 \times \mu \times d_{kv}$ parameters. These embedding matrices are shared in each encoder layer and each head.  Here $\mu = 51$ is the total number of relations and $d_{kv}$ is the dimension of the key/value states in self-attention (64 in T5-small/base/large and 128 in T5-3B). The overall increase of parameters is less than 0.01\%.

\begin{table}[h]
\centering
\begin{tabular}{ll}
\toprule
Approach      & \#param                       \\ 
\midrule
T5-small      & 60,506,624                      \\
RASAT(-small) & \textbf{60,512,768(+0.0107\%)}  \\ 
\midrule
T5-base       & 222,903,552                     \\
RASAT(-base)  & \textbf{222,909,696(+0.0029\%)} \\ 
\midrule
T5-large      & 737,668,096                     \\
RASAT(-large) & \textbf{737,674,240(+0.0009\%)} \\ 
\midrule
T5-3B         & 2,851,598,336                    \\
RASAT(-3B)    & \textbf{2,851,610,624(+0.0005\%)} \\ 
\bottomrule
\end{tabular}
\caption{The number of paramerter comparation between RASAT and the same size T5 model.}
\label{tab:param}
\end{table}

\begin{table*}[!ht]
\centering
\resizebox{2.0\columnwidth}{!}{
\begin{tabular}{ll}
\toprule
Question & What   is the average, minimum, and maximum age of all singers from France?                   \\ \midrule
 Tree-based model      & \textbf{\textcolor{blue}{SELECT}}   \textbf{\textcolor{blue}{AVG}}(singer.Age) , \textbf{\textcolor{blue}{MAX}}(singer.Age) , \textbf{\textcolor{blue}{MIN}}(singer.Age) \textbf{\textcolor{blue}{FROM}} singer \textbf{\textcolor{blue}{WHERE}} singer.Country = \boxed{$"$value$"$}  \\
 RASAT     & \textbf{\textcolor{blue}{SELECT}}   \textbf{\textcolor{blue}{AVG}}(singer.Age) , \textbf{\textcolor{blue}{MAX}}(singer.Age) , \textbf{\textcolor{blue}{MIN}}(singer.Age) \textbf{\textcolor{blue}{FROM}} singer \textbf{\textcolor{blue}{WHERE}} singer.Country = \boxed{$"$France$"$} \\ \bottomrule
\end{tabular}}
\caption{An example to show the difference between AST-based decoder model's output and T5's output.}
\label{tab:output_compare}
\end{table*}

\begin{table*}[!t]
\centering
\resizebox{2.0\columnwidth}{!}{
\begin{tabular}{ll}
\hline
\toprule
Description  & A database about   employee hiring and evaluation.                                              \\
Goal         & \textbf{Find   cities which more than one employee under age 30 come from.}                              \\
\midrule
Question \#1 & Find   all employees who are under age 30.                                                      \\
T5-3B        & \textbf{\textcolor{blue}{SELECT}} * \textbf{\textcolor{blue}{FROM}} employee   \textbf{\textcolor{blue}{WHERE}} age \textless 30                                                 \\
RASAT        & \textbf{\textcolor{blue}{SELECT}} * \textbf{\textcolor{blue}{FROM}} employee   \textbf{\textcolor{blue}{WHERE}} age \textless 30                                                 \\ \midrule
Question \#2 & Which   cities did they come from?                                                              \\
T5-3B        & \textbf{\textcolor{blue}{SELECT}} city \textbf{\textcolor{blue}{FROM}}   employee \textbf{\textcolor{blue}{WHERE}} age \textless 30                                              \\
RASAT        & \textbf{\textcolor{blue}{SELECT}} city \textbf{\textcolor{blue}{FROM}}   employee \textbf{\textcolor{blue}{WHERE}} age \textless 30                                              \\ \midrule
Question \#3 & Show   the cities \underline{from which} more than one employee originated.                                 \\
T5-3B        & \textbf{\textcolor{blue}{SELECT}} city \textbf{\textcolor{blue}{FROM}}   employee \textbf{\textcolor{blue}{GROUP BY}} city \textbf{\textcolor{blue}{HAVING COUNT}}(*) \textgreater 1                        \\
RASAT        & \textbf{\textcolor{blue}{SELECT}} city \textbf{\textcolor{blue}{FROM}}   employee \boxed{\textbf{\textcolor{blue}{WHERE} \textcolor{red}{age \textless 30}}} \textbf{\textcolor{blue}{GROUP BY}} city \textbf{\textcolor{blue}{HAVING COUNT}}(*) \textgreater 1 \\ 
\midrule

\toprule
Description  & A database about   courses and teachers.                                                                                                                                             \\
Goal         & \makecell[l]{\textbf{Show   names of teachers and the courses they are arranged to teach in ascending   alphabetical} \\ \textbf{order of the teacher's name.}}                                                         \\ 
\midrule
Question \#1 & Find   all the course arrangements.                                                                                                                                                  \\
T5-3B        & \textbf{\textcolor{blue}{SELECT}} * \textbf{\textcolor{blue}{FROM}}   course\_arrange                                                                                                                                                      \\
RASAT        & \textbf{\textcolor{blue}{SELECT}} * \textbf{\textcolor{blue}{FROM}}   course\_arrange                                                                                                                                                      \\ 
\midrule
Question \#2 & Show   names of teachers and the courses they are arranged to teach.                                                                                                                 \\
T5-3B        & \textbf{\textcolor{blue}{SELECT}} T2.name,   T1.course \textbf{\textcolor{blue}{FROM}} course\_arrange \textbf{\textcolor{blue}{AS}} T1 \textbf{\textcolor{blue}{JOIN}} teacher \textbf{\textcolor{blue}{AS}} T2 \textbf{\textcolor{blue}{ON}} T1.teacher\_id =   T2.teacher\_id                                                                       \\
RASAT        & \makecell[l]{\textbf{\textcolor{blue}{SELECT}} T2.name,   T3.course \textbf{\textcolor{blue}{FROM}} course\_arrange \textbf{\textcolor{blue}{AS}} T1 \textbf{\textcolor{blue}{JOIN}} teacher \textbf{\textcolor{blue}{AS}} T2 \textbf{\textcolor{blue}{ON}} T1.teacher\_id =   T2.teacher\_id \\ \boxed{\textbf{\textcolor{blue}{JOIN} \textcolor{red}{course} \textbf{\textcolor{blue}{AS}} \textcolor{red}{T3} \textbf{\textcolor{blue}{ON}} \textcolor{red}{T1.course\_id = T3.course\_id}}}}                    \\ \midrule
Question \#3 & Sort   the results by teacher's name                                                                                                                                                 \\
T5-3B        & \makecell[l]{\textbf{\textcolor{blue}{SELECT}} T2.name,   T1.course \textbf{\textcolor{blue}{FROM}} course\_arrange \textbf{\textcolor{blue}{AS}} T1 \textbf{\textcolor{blue}{JOIN}} teacher \textbf{\textcolor{blue}{AS}} T2 \textbf{\textcolor{blue}{ON}} T1.teacher\_id =   T2.teacher\_id \\ \textbf{\textcolor{blue}{ORDER BY}} T2.name}                                                      \\
RASAT        & \makecell[l]{\textbf{\textcolor{blue}{SELECT}} T3.name,   T2.course \textbf{\textcolor{blue}{FROM}} course\_arrange \textbf{\textcolor{blue}{AS}} T1 \textbf{\textcolor{blue}{JOIN}} course \textbf{\textcolor{blue}{AS}} T2 \textbf{\textcolor{blue}{ON}} T1.course\_id =   T2.course\_id\\ \boxed{\textbf{\textcolor{blue}{JOIN} \textcolor{red}{teacher} \textbf{\textcolor{blue}{AS}} \textcolor{red}{T3} \textbf{\textcolor{blue}{ON}} \textcolor{red}{T1.teacher\_id = T3.teacher\_id}}} \textbf{\textcolor{blue}{ORDER BY}}   T3.name} \\ \bottomrule
\end{tabular}
}
\caption{Some examples in the SParC dev set. RASAT gives all correct predictions in these cases while the original T5-3B model fails.  }
\label{tab:sparc_case_study}
\end{table*}

\section{Output Comparation between T5 and Tree-based Decoder Model}
Here we show the output difference between models used AST-tree-based decoder and T5. As it shown in Table \ref{tab:output_compare}, models used AST-tree-based decoder (such as RAT-SQL \citep{wang-etal-2020-rat}, LGESQL \citep{cao-etal-2021-lgesql}) usually use a place holder (i.e. "value") to represent the real value("France" in this example). These output can not be executed in real database and they fail to evalute in EXecution Accuracy(EX/QEX/IEX) metric.

\section{Case Study}
\label{sec:appendix_case_study}

In Table \ref{tab:sparc_case_study}, we demonstrate how the introduced relation could help the model predict SQL structures more accurately by demonstrating 2 examples of question-SQL pairs sampled from the SParC dev set. We compare the predictions from T5-3B and our model, and both the two examples have three turns in the interaction. For the first case, the vanilla T5-3B model neglects the condition "employees who are under age 30" when answering Question \#3, while RASAT-SQL predicts it correctly by exploiting the relations inside the contexts. For the second case, the database schema is more complex, and the table \texttt{course\_arrange} has no such a column called \texttt{course}. If one would like to access column \texttt{course}, the foreign key must be used. RASAT gives the correct SQL since these types of relational structures are explicitly embedded in the RASAT model, while the vanilla T5-3B fails to do it.

\section{Relations Used in Experiment}
\label{sec:appendix_relation}
Table \ref{tab:all_relations} shows all relations used in our experiment while most of these are consistant with RAT-SQL \citep{wang-etal-2020-rat} and LGESQL \citep{cao-etal-2021-lgesql}. There are total 51 kinds relation used.

\begin{table*}
    \centering
    \resizebox{2.0\columnwidth}{!}{
        \begin{tabular}{llll} 
    \toprule
    Head H            & Tail T            & Edge label                   & Description                                                                        \\ 
    \midrule
    \multirow{8}{*}{$\mathcal{Q}$} & \multirow{8}{*}{$\mathcal{Q}$} & Question-Question-Dist*     &
    Question item H is at a distance of * before question item T in the input question \\
    &                     & Question-Question-Identity     &        Question item H is question item T itself                                                                          \\
    &                     & Question-Question-Generic    &   Question item H and question item T has no pre-defined relation \\                                                                                 
    \cmidrule{3-4} 
    &                     & Forward-Syntax               & \multirow{3}{*}{Question item H has a forward/reverse/no syntactic dependencies on question item T}                    \\
    &                     & Backward-Syntax              &                                                                                     \\
    &                     & None-Syntax                  &                                                                                    \\ 
    \cmidrule{3-4} 
    &                     & Co\_Relations                &  Question item H and question item T are considered as a whole in coreference relation                           \\
    &                     & Coref\_Relations             & Question item H is the coreference of question item T                                                                              \\ \midrule
   $\mathcal{Q}$                  &$\mathcal{S}$                  & Question-*-Generic           & Question item H and database item T has no pre-defined relation                   \\ \midrule
    \multirow{3}{*}{$\mathcal{Q}$}  & \multirow{3}{*}{$\mathcal{T}$}  & Question-Table-Exactmatch    & \multirow{3}{*}{Question item H is spelled exactly/partially/not the same as table item T}                                  \\
                        &                     & Question-Table-Partialmatch  &\\
                        &                     & Question-Table-Nomatch       &   
                        \\ \midrule
    \multirow{4}{*}{$\mathcal{Q}$}  & \multirow{4}{*}{C}  & Question-Column-Exactmatch   & \multirow{3}{*}{Question item H is spelled exactly/partially/not the same as column item T}                                  \\
                        &                     & Question-Column-Partialmatch &                                                                                    \\
                        &                     & Question-Column-Nomatch      &                                                                                    \\ \cmidrule{3-4} 
                        &                     & Question-Column-Valuematch   &         Question item H is spelled exactly the same as a value in column item T                                                                          \\ \midrule
   $\mathcal{S}$                  &$\mathcal{Q}$                  & *-Question-Generic           & Database item H and question item T has no pre-defined relation                                                      \\ \midrule
   $\mathcal{S}$                  &$\mathcal{S}$                  & *-*-Identity                 & Database item H is database item T itself                                                        \\ \midrule
   $\mathcal{S}$                  &$\mathcal{T}$                  & *-Table-Generic              & Database item H and table item T has no pre-defined relation                                                           \\ \midrule
   $\mathcal{S}$                  &$\mathcal{C}$                  & *-Column-Generic             & Database item H and column item T has no pre-defined relation                                                           \\ \midrule
    \multirow{3}{*}{$\mathcal{T}$}  & \multirow{3}{*}{$\mathcal{Q}$}  & Table-Question-Exactmatch    & \multirow{3}{*}{Table item H is spelled exactly/partially/not the same as question item T}                                      \\
                        &                     & Table-Question-Partialmatch  &                                                                                    \\
                        &                     & Table-Question-Nomatch       &                                                                                    \\ \midrule
   $\mathcal{T}$                  &$\mathcal{S}$                  & Table-*-Generic              & Table item H and database item T has no pre-defined relation                                                             \\ \midrule
    \multirow{5}{*}{$\mathcal{T}$}  & \multirow{5}{*}{$\mathcal{T}$}  & Table-Table-Generic          & Table item H and table item T has no pre-defined relation                                \\ \cmidrule{3-4} 
                        &                     & Table-Table-Identity         & Table item H is table item T itself                                                                \\ \cmidrule{3-4} 
                        &                     & Table-Table-Fk               & At least one column in table item H is a foreign key for certain column in table item T     \\
                        &                     & Table-Table-Fkr              &       At least one column in table item T is a foreign key for certain column in table item H                                                                             \\
                        &                     & Table-Table-Fkb              &   Table item H and  T satisfy both "Table-Table-Fk" and "Table-Table-Fkr" relations                                                                        \\ \midrule
    \multirow{3}{*}{$\mathcal{T}$}  & \multirow{3}{*}{$\mathcal{C}$}  & Table-Column-Pk              & Column item T is the primary key for table item H                                            \\  
                        &                     & Table-Column-Has             & Column item T belongs to table item H                                                                \\ \cmidrule{3-4} 
                        &                     & Table-Column-Generic         & Table item H and column item T has no pre-defined relation                       \\ \midrule
    \multirow{4}{*}{$\mathcal{C}$}  & \multirow{4}{*}{$\mathcal{Q}$}  & Column-Question-Exactmatch    & \multirow{3}{*}{Column item H is spelled exactly/partially/not the same as table item T}                                     \\
                        &                     & Column-Question-Partialmatch &                                                                                    \\
                        &                     & Column-Question-Nomatch      &                                                                                    \\\cmidrule{3-4} 
                        &                     & Column-Question-Valuematch   &    Column item H is spelled exactly the same as a value in question item T                                                                                  \\ \midrule
   $\mathcal{C}$                  &$\mathcal{S}$                  & Column-*-Generic             & Column item H and database item T has no pre-defined relation                                                           \\ \midrule
    \multirow{3}{*}{$\mathcal{C}$}  & \multirow{3}{*}{$\mathcal{T}$}  & Column-Table-Pk              & Column item H is the primary key for table item T                               \\
                        &                     & Column-Table-Has             &      Column item H belongs to table item T                                                                               \\
                        &                     & Column-Table-Generic         &         Column item H and table item T has no pre-defined relation                                                                           \\ \midrule
    \multirow{5}{*}{$\mathcal{C}$}  & \multirow{5}{*}{$\mathcal{C}$}  & Column-Column-Identity       & Column item H is column item T itself                                                            \\ \cmidrule{3-4} 
                        &                     & Column-Column-Sametable      & Column item H and column item T are in the  same table                                                              \\ \cmidrule{3-4} 
                        &                     & Column-Column-Fk             & \multirow{2}{*}{Column item H has a forward/reverse foreign key constraint relation with Column item T}                                              \\
                        &                     & Column-Column-Fkr            &                                                                                    \\ \cmidrule{3-4} 
                        &                     & Column-Column-Generic        & Column item H and column item T has no pre-defined relation                      \\ \midrule
   $\mathcal{C}$                  &$\mathcal{V}$                  & Has-Dbcontent                & Db content item T belongs to column item H                                                                      \\ \midrule
   $\mathcal{V}$                  &$\mathcal{C}$                  & Has-Dbcontent-R              & Db content item H belongs to column item T                                                                       \\ \midrule
                  &               & No-Relation                  & Item H and  item T has no relation (Used when item H or item T is a delimiter)                                                        \\ 
    \bottomrule
    \end{tabular}
    }
\caption{All relations used in our experiment. $\mathcal{V}$ is the matched question item that extracted from $\mathcal{Q}$. }
\label{tab:all_relations}
\end{table*}

\end{document}